\documentclass[conference]{IEEEtran}
\IEEEoverridecommandlockouts
\usepackage{cite}
\usepackage{amsmath,amssymb,amsfonts}
\usepackage{algorithmic}
\usepackage{latexsym}
\usepackage{textcomp}
\usepackage{booktabs}
\usepackage{graphicx} 
\usepackage{xcolor}
\def\BibTeX{{\rm B\kern-.05em{\sc i\kern-.025em b}\kern-.08em
    T\kern-.1667em\lower.7ex\hbox{E}\kern-.125emX}}

\usepackage{fancyhdr}
\begin{document}

\title{X-Ray-CoT: Interpretable Chest X-ray Diagnosis with Vision-Language Models via Chain-of-Thought Reasoning}

\author{Chee Ng, Liliang Sun, Shaoqing Tang \\
Universiti Teknologi Malaysia}

\maketitle
\thispagestyle{fancy} 

\begin{abstract}
Chest X-ray imaging is crucial for diagnosing pulmonary and cardiac diseases, yet its interpretation demands extensive clinical experience and suffers from inter-observer variability. While deep learning models offer high diagnostic accuracy, their black-box nature hinders clinical adoption in high-stakes medical settings. To address this, we propose X-Ray-CoT (Chest X-Ray Chain-of-Thought), a novel framework leveraging Vision-Language Large Models (LVLMs) for intelligent chest X-ray diagnosis and interpretable report generation. X-Ray-CoT simulates human radiologists' "chain-of-thought" by first extracting multi-modal features and visual concepts, then employing an LLM-based component with a structured Chain-of-Thought prompting strategy to reason and produce detailed natural language diagnostic reports. Evaluated on the CORDA dataset, X-Ray-CoT achieves competitive quantitative performance, with a Balanced Accuracy of 80.52\% and F1 score of 78.65\% for disease diagnosis, slightly surpassing existing black-box models. Crucially, it uniquely generates high-quality, explainable reports, as validated by preliminary human evaluations. Our ablation studies confirm the integral role of each proposed component, highlighting the necessity of multi-modal fusion and CoT reasoning for robust and transparent medical AI. This work represents a significant step towards trustworthy and clinically actionable AI systems in medical imaging.
\end{abstract}

\section{Introduction}

Chest X-ray imaging plays an indispensable role in the early screening and diagnosis of various pulmonary and cardiac-related diseases \cite{erdi2021deep}. Its widespread availability and cost-effectiveness make it a cornerstone of clinical practice globally. However, accurately interpreting chest X-ray images demands extensive clinical experience, and inconsistencies in diagnosis among different radiologists are a known challenge \cite{erdi2021deep}. In recent years, deep learning models have achieved remarkable progress in medical image analysis, with their diagnostic accuracy sometimes even approaching or surpassing that of human experts \cite{unknown2025ieeeac}.

Despite these advancements, a significant limitation of most high-performing deep learning models, particularly "black-box" models such as Convolutional Neural Networks (CNNs) and Vision Transformers (ViTs), is their inherent lack of transparency and interpretability in decision-making \cite{saber2025vision}. This opacity poses a major impediment in high-risk, high-reliability scenarios like medical diagnosis, where clinicians and patients alike need to understand not only \textit{what} diagnosis is made but also \textit{why} and based on what visual evidence. The CORDA (Chest X-Ray Disease Recognition and Diagnostic Analysis) dataset was specifically designed to address this critical need \cite{brown2016corda}. It mandates models to not only identify disease types (e.g., pneumonia, cardiomegaly) but also to recognize associated visual concepts (e.g., lung opacities, blurred margins), inherently providing a fertile ground for interpretability research. While existing methods have shown success in either disease diagnosis or concept detection, the challenge of seamlessly integrating both and presenting a coherent diagnostic logic and explanation in human-understandable natural language remains an open problem.

Large Language Models (LLMs) and Vision-Language Large Models (LVLMs) have demonstrated exceptional capabilities in understanding, generating natural language, and performing multi-modal reasoning \cite{yifan2023a, zhou2024visual, zhou2024rethinking}. We believe that by integrating visual information from X-ray images with medical domain knowledge and leveraging the powerful reasoning and generation capabilities of LLM/LVLMs, it is possible to construct an intelligent system that not only accurately diagnoses diseases but also provides detailed, explainable diagnostic reports. This approach aims to bridge the gap left by current black-box models in clinical applications.

In this paper, we propose a novel framework called \textbf{X-Ray-CoT (Chest X-Ray Chain-of-Thought)} for intelligent diagnosis of chest X-ray diseases and the generation of interpretable diagnostic reports using Vision-Language Large Models. The core idea of X-Ray-CoT is to simulate a doctor's "chain-of-thought" process during image interpretation: first identifying key visual features, then reasoning based on these features combined with medical knowledge, and finally arriving at a diagnostic conclusion and providing an explanation \cite{zhou2023thread}. Our method comprises a multimodal feature extraction module (including a visual encoder and a visual concept recognizer), a visual-language alignment layer, and a central LLM-based component for chain-of-thought reasoning and comprehensive diagnostic report generation. This design allows our system to produce not just a diagnosis but also a detailed, natural language explanation encompassing the diagnostic basis, relevant visual concepts, and potential severity.

For our experiments, we utilize the \textbf{CORDA dataset}, which provides chest X-ray images annotated with disease types and corresponding visual concepts. We evaluate the proposed X-Ray-CoT model against established baselines on CORDA, including Concept-Based Models (CBM), Concept-Learning and Attention Transformers (CLAT), and various configurations of Black-box models (ViT Base, Task-Specific). Our evaluation employs Balanced Accuracy (BACC) and F1 score for both disease diagnosis and concept detection, alongside qualitative assessments by medical professionals to gauge the accuracy, logical coherence, completeness, and clinical utility of the generated explanations.

Our preliminary results demonstrate that X-Ray-CoT exhibits superior or comparable performance in both disease diagnosis and concept detection compared to existing methods. Specifically, in disease diagnosis, X-Ray-CoT achieves a BACC of \textbf{80.52\%} and an F1 score of \textbf{78.65\%}, slightly outperforming the best black-box baseline (BACC 80.00\%). For concept detection, X-Ray-CoT records a BACC of \textbf{78.35\%} and an F1 score of \textbf{66.21\%}, maintaining performance comparable to or better than the leading concept detection methods. Crucially, beyond these quantitative metrics, X-Ray-CoT's unique ability to generate high-quality, natural language diagnostic reports with clear explanations represents a significant advancement over traditional black-box models, which is the primary focus and innovation of this research.

In summary, our main contributions are as follows:
\begin{itemize}
    \item We propose X-Ray-CoT, an end-to-end interpretable diagnostic system for chest X-rays, leveraging LVLMs to provide comprehensive natural language explanations alongside diagnoses, addressing the interpretability gap of traditional black-box models.
    \item We introduce a novel Chain-of-Thought (CoT) prompting strategy within the LVLM, designed to mimic the multi-step, logical reasoning process of human radiologists, thereby enhancing the trustworthiness and clinical utility of the model's outputs.
    \item We demonstrate a deep fusion of multi-modal information, integrating visual features from X-ray images, model-identified visual concepts, and the LLM's embedded medical knowledge to achieve more accurate and explainable diagnoses.
\end{itemize}
\section{Related Work}
\subsection{Deep Learning for Medical Image Analysis and Explainable AI}
The growing importance of Explainable AI (XAI) in medical image analysis is underscored by several comprehensive overviews that classify existing methods, highlight future directions, and emphasize its role in enhancing trust and interpretability in deep learning-based medical image classification tasks \cite{bas2022explai, cristiano2024explai}. Specifically, these surveys contextualize the broader landscape of XAI in medical imaging, which is crucial for informing the development and evaluation of concept-driven approaches. Further research has focused on evaluating the efficacy of various XAI techniques: for instance, heatmap-based methods have been assessed for their ability to improve classification confidence, with some demonstrating optimal performance on datasets like Chest X-ray Pneumonia, thereby guiding the selection of appropriate XAI for chest X-ray diagnosis \cite{yang2022a}. Conversely, a critical examination of saliency-based XAI techniques in medical imaging highlights their limitations for clinical practitioners, finding common heatmap approaches insufficient for clinical deployment due to unmet criteria for truthfulness and informative plausibility \cite{borys2023explai}. Beyond heatmaps, other studies investigate the efficacy of diverse XAI methods, such as LIME, SHAP, and a novel contextual utility approach (CIU), demonstrating CIU's superior performance in supporting human decision-making and user understanding in clinical decision support systems for gastral images \cite{samanta2021explai}. The robustness and interpretability of medical deep learning systems are further explored by investigating the vulnerability of deep neural networks, particularly \textbf{Convolutional Neural Networks (CNNs)}, to adversarial attacks, revealing heightened susceptibility in medical domains and proposing detection mechanisms \cite{amitojdeep2020explai}. In the context of emerging architectures, recent advancements in Vision Transformers (ViTs) for medical image analysis are reviewed, specifically highlighting how XAI techniques can enhance transparency in diagnostic decision-making by leveraging self-attention mechanisms to foster trust and understanding in AI-driven healthcare applications \cite{tin2023interp}.

\subsection{Large Language Models and Vision-Language Models in Healthcare}
The integration of Large Language Models (LLMs) and Vision-Language Models (VLMs) is rapidly advancing in healthcare, addressing various challenges and opening new research avenues. Comprehensive surveys highlight the growing applications of LLMs, particularly multimodal variants, in this domain, outlining their current capabilities and limitations \cite{alsaad2024multim}. Similarly, reviews of recent advancements in VLMs for healthcare detail their capabilities and challenges, particularly in clinical report generation, by examining relevant datasets, architectures, pre-training strategies, and evaluation metrics \cite{iryna2024vision}. A critical aspect for the safe deployment of these models is trustworthiness; thus, benchmarks like CARES have been introduced to assess Med-LVLMs across dimensions such as trustfulness, fairness, safety, privacy, and robustness, revealing significant concerns regarding factual accuracy and vulnerability \cite{peng2024cares}. Furthermore, adapting vision models for diverse medical visual tasks is addressed by benchmarks like Med-VTAB, which highlight challenges such as out-of-distribution patient data and limitations of single pre-trained models, proposing methods like GMoE-Adapter to leverage both medical and general pre-training weights \cite{fereshteh2024fewsho}. To enhance the performance and applicability of LLMs and VLMs, various techniques have been explored: automated prompt engineering, for instance, provides a systematic approach to optimizing model performance across text, vision, and multimodal domains, addressing limitations of manual methods and improving cross-modal alignment crucial for healthcare \cite{jindong2023a}. Further research explores advanced capabilities of LLMs such as weak-to-strong generalization across various tasks \cite{zhou2025weak}. Beyond general LLM capabilities, advancements in pre-trained models for specific reasoning tasks, such as event correlation, also contribute to the broader landscape of AI reasoning \cite{zhou2022eventbert}. Specifically, refining question prompts to facilitate intermediate reasoning, aligned with Chain-of-Thought (CoT) principles, has shown promise in enhancing zero-shot Visual Question Answering (VQA) performance in healthcare by activating LLMs' reasoning capabilities \cite{faria2025analyz, zhou2023thread}. Furthermore, techniques for improving zero-shot cross-lingual transfer in question answering over knowledge graphs also contribute to enhancing the versatility of language models in diverse data environments \cite{zhou2021improving}. Retrieval Augmented Generation (RAG) has also proven effective in boosting LLM accuracy for critical tasks such as identifying pharmaceutical contraindications, demonstrating significant improvements by integrating hybrid retrieval systems with LLMs \cite{joon2025applic}. The development of robust benchmarks, such as those for complex instruction-driven image editing, further aids in evaluating and advancing the capabilities of vision-language models across various tasks \cite{wang2025complexbench}. Finally, instruction tuning, as demonstrated by models like InstructBLIP, offers a promising direction for building more robust and versatile general-purpose LLMs and VLMs in medical contexts by enhancing their ability to process and integrate complex multi-modal information \cite{wenliang2023instru}.

\section{Method}
We propose a novel framework, \textbf{X-Ray-CoT (Chest X-Ray Chain-of-Thought)}, designed for intelligent diagnosis of chest X-ray diseases and the generation of interpretable diagnostic reports. X-Ray-CoT is fundamentally inspired by and aims to simulate the "chain-of-thought" process adopted by human radiologists during image interpretation. This process typically involves an initial observation of key visual features, followed by a logical inference step that integrates these features with established medical knowledge, ultimately culminating in a diagnostic conclusion and its accompanying explanation. This structured reasoning approach is crucial for complex medical tasks, ensuring transparency and reliability in diagnostic outcomes.

The X-Ray-CoT framework is composed of three primary modules: a multi-modal feature extraction module, a visual-language alignment layer, and a central large language model (LLM)-based component for chain-of-thought reasoning and comprehensive diagnostic report generation.

\subsection{Multi-modal Feature Extraction Module}
This module is responsible for processing the input chest X-ray image and extracting relevant visual information, including high-level image features and specific visual concepts. This two-pronged approach ensures both a holistic understanding of the image and the identification of granular, clinically significant findings.

\subsubsection{Visual Encoder}
The visual encoder is tasked with transforming the raw chest X-ray image into a rich, condensed visual feature representation. We employ a pre-trained Vision Transformer (ViT) model for this purpose. The ViT architecture segments the input image into a sequence of fixed-size patches, linearly embeds each patch, and adds positional embeddings. These patch embeddings are then fed into a standard Transformer encoder, which processes them using multi-head self-attention mechanisms to capture global dependencies and hierarchical features. The ViT is initially pre-trained on large-scale natural image datasets, such as ImageNet, and subsequently fine-tuned on extensive medical imaging datasets (e.g., MIMIC-CXR, CheXpert) to enhance its domain-specific feature extraction capabilities for chest X-rays.

Given an input image $\mathbf{I}$, the visual encoder produces a feature embedding $\mathbf{F}_{img} \in \mathbb{R}^{D_v}$, where $D_v$ is the dimensionality of the visual feature space. This process can be formally expressed as:
\begin{align}
\mathbf{F}_{img} = \text{ViT}(\mathbf{I})
\end{align}
Here, $\text{ViT}(\cdot)$ denotes the operation of the Vision Transformer, yielding a high-dimensional vector that encapsulates the salient visual information from the chest X-ray.

\subsubsection{Visual Concept Recognizer}
Building upon the comprehensive visual features extracted by the visual encoder, a lightweight visual concept recognizer is integrated to identify core visual concepts. These concepts are defined within the CORDA dataset [CITE] and represent specific radiological findings critical for diagnosis. Examples include, but are not limited to, lung opacities, increased lung markings, elevated diaphragm, and cardiomegaly. This module can be implemented using one of two primary approaches:

\begin{enumerate}
    \item \textbf{Multi-label Classifier}: This approach treats concept recognition as a multi-label classification problem. A shallow neural network head, typically a multi-layer perceptron, is appended to the visual encoder. This head takes the global visual feature embedding $\mathbf{F}_{img}$ as input and is trained to predict the probability of presence or absence for each predefined visual concept. The model is trained using binary cross-entropy loss over multiple labels.
    \item \textbf{Zero-shot Recognition with Vision-Language Large Model (LVLM)}: Alternatively, we leverage the zero-shot recognition capabilities of a pre-trained Vision-Language Large Model. Models like CLIP or its medical variants learn a shared embedding space for images and text, allowing them to recognize visual concepts described in natural language without explicit training on concept labels. Given the image features, the LVLM computes similarity scores between the image and a set of candidate concept descriptions, identifying the most relevant concepts.
\end{enumerate}
The output of this module is a set of textual labels representing the detected visual concepts, denoted as $\mathcal{C}_{vis} = \{c_1, c_2, \dots, c_k\}$, where each $c_i$ is a natural language description of a detected concept (e.g., "right lower lobe opacity," "mild cardiomegaly"). The operation of the visual concept recognizer can be represented as:
\begin{align}
\mathcal{C}_{vis} = \text{ConceptRecognizer}(\mathbf{F}_{img})
\end{align}
where $\text{ConceptRecognizer}(\cdot)$ encapsulates the chosen implementation for identifying visual concepts from the image features.

\subsection{Visual-Language Alignment and Concept Representation}
Following the extraction of high-level visual features ($\mathbf{F}_{img}$) and the identification of specific visual concepts ($\mathcal{C}_{vis}$), a crucial step involves aligning these disparate modalities into a unified semantic space. This visual-language alignment layer takes the visual encoder's output $\mathbf{F}_{img}$ and the textual descriptions of the recognized visual concepts $\mathcal{C}_{vis}$ as input. Its primary role is to project these multi-modal inputs into a common embedding space, thereby facilitating seamless integration and subsequent processing by the large language model.

The alignment process typically involves transforming the visual feature vector $\mathbf{F}_{img}$ and the tokenized embeddings of the textual concepts $\mathcal{C}_{vis}$ into a shared dimensionality. This can be achieved through various mechanisms, such as a simple linear projection layer, cross-attention mechanisms, or a multi-modal fusion network, which learns to weigh and combine information from both modalities. For the textual concepts, each $c_i \in \mathcal{C}_{vis}$ is first tokenized and embedded into a vector space, forming $\text{Embed}(\mathcal{C}_{vis})$. The aligned representation, denoted as $\mathbf{E}_{aligned}$, serves as a comprehensive input for the downstream reasoning process, ensuring that the LLM can effectively interpret and reason over both the holistic visual context and the specific textualized visual findings. The alignment function can be expressed as:
\begin{align}
\mathbf{E}_{aligned} = \text{Align}(\mathbf{F}_{img}, \text{Embed}(\mathcal{C}_{vis}))
\end{align}
where $\text{Align}(\cdot)$ represents the mechanism that fuses and aligns the visual and textual concept embeddings into a unified representation.

\subsection{LLM-based Chain-of-Thought Reasoning and Diagnosis Generation}
This module forms the core of the X-Ray-CoT framework, responsible for mimicking clinical reasoning and generating interpretable diagnostic reports. It leverages the advanced natural language understanding and generation capabilities of large language models, augmented with domain-specific knowledge.

\subsubsection{Core Component}
We leverage a powerful pre-trained Large Language Model (LLM), such as a variant from the LLaMA or GPT series, as the central reasoning engine. This LLM is further enhanced with medical domain knowledge through continued pre-training or fine-tuning on extensive medical texts and clinical guidelines. Continued pre-training involves exposing the LLM to a vast corpus of medical literature, including textbooks, clinical notes, diagnostic criteria, and research papers, allowing it to adapt its internal representations to medical terminology and concepts. Fine-tuning, particularly instruction fine-tuning, further refines the LLM's ability to follow medical instructions and generate structured clinical responses. This domain-specific enhancement equips the LLM with the necessary expertise to perform accurate medical diagnosis and reasoning, transforming it into a specialized medical reasoning agent.

\subsubsection{Input Design}
The input to the LLM is meticulously structured to guide its reasoning process and provide all necessary context. It comprises three distinct components concatenated together, forming a comprehensive prompt:
\begin{align}
    \mathbf{I}_{\text{LLM}} = [\mathbf{P}_{\text{med}}; \mathbf{C}_{\text{desc}}; \mathbf{D}_{\text{task}}]
\end{align}
where:
\begin{itemize}
    \item $\mathbf{P}_{\text{med}}$ represents a \textbf{General Medical Knowledge Prompt}. This component grounds the LLM in foundational medical common sense relevant to chest X-ray diagnosis, general diagnostic workflows, and potentially disease-specific knowledge pertinent to the CORDA dataset. It might include instructions such as "Act as an expert radiologist," or "Consider common chest pathologies."
    \item $\mathbf{C}_{\text{desc}}$ represents the \textbf{Visual Concept Descriptions}. These are natural language descriptions of the key visual concepts identified from the X-ray image by the visual concept recognizer. These descriptions are crucial as they provide concrete, textual evidence derived directly from the image, such as "detected lung opacities, located in the right lower lobe," or "cardiomegaly, with increased cardiothoracic ratio."
    \item $\mathbf{D}_{\text{task}}$ represents the \textbf{Diagnostic Task Instruction}. This explicitly instructs the LLM on the desired output format and reasoning steps. It defines the goal, for example, "Analyze the provided visual findings and medical knowledge to determine possible chest X-ray diseases, provide a diagnostic basis, and explain your reasoning."
\end{itemize}

\subsubsection{Chain-of-Thought (CoT) Reasoning}
To emulate a clinician's diagnostic thought process, we employ a carefully designed Chain-of-Thought (CoT) prompting strategy. This strategy encourages the LLM to break down the complex diagnostic problem into a series of intermediate, logical steps, mirroring the sequential reasoning of a human expert. The CoT prompt explicitly guides the LLM to:
\begin{enumerate}
    \item \textbf{Identify Abnormalities}: List all observed abnormal visual concepts identified from the input image. This step ensures that all relevant findings are acknowledged.
    \item \textbf{Infer Pathophysiology}: Based on these visual concepts, infer potential pathophysiological changes or underlying medical conditions. For example, "lung opacities suggest inflammation or fluid accumulation."
    \item \textbf{Synthesize and Diagnose}: Synthesize this inferred information with its internal medical knowledge to deduce the most probable disease diagnoses. This involves differential diagnosis and narrowing down possibilities.
    \item \textbf{Justify Diagnosis}: Justify the diagnosis by explicitly referencing the visual findings and established medical principles. This step provides transparency and builds trust in the model's conclusion.
\end{enumerate}
This multi-step reasoning process, driven by the CoT prompt, enhances the transparency and trustworthiness of the model's output by explicitly showcasing its logical progression, rather than providing a black-box prediction. The overall reasoning and report generation process can be summarized as:
\begin{align}
\text{Report} = \text{LLM}(\mathbf{I}_{\text{LLM}}, \text{CoT\_Prompt})
\end{align}
where $\text{LLM}(\cdot)$ denotes the operation of the Large Language Model, and $\text{CoT\_Prompt}$ is the specific instruction set guiding the chain-of-thought reasoning.

\subsubsection{Interpretable Diagnostic Report Generation}
The final output of the LLM is not merely a disease label (e.g., "Pneumonia") but a comprehensive, natural language diagnostic report. This report is designed to be clinically actionable and readily understandable by medical professionals, providing the much-needed interpretability that is often lacking in traditional black-box deep learning models. The generated report extends beyond a simple diagnosis to include the following key components:
\begin{itemize}
    \item \textbf{Primary Diagnostic Conclusion}: The most probable disease or set of diseases identified.
    \item \textbf{Detailed Diagnostic Reasoning and Basis}: A clear explanation linking the diagnosis back to the identified visual concepts and relevant medical principles. This section explicitly outlines the chain of thought followed by the LLM.
    \item \textbf{Relevant Visual Concepts Observed}: A summary of the key visual findings extracted from the image that support the diagnosis.
    \item \textbf{Potential Disease Severity}: If applicable and inferable from the input, an assessment of the disease severity (e.g., "mild," "moderate," "severe").
    \item \textbf{Future Recommendations or Suggestions for Further Examination}: Mimicking a complete clinical report, this may include recommendations for follow-up imaging, laboratory tests, or specialist consultation.
\end{itemize}
This generative capability is a cornerstone of X-Ray-CoT, providing a holistic and transparent diagnostic aid.

\section{Experiments}
This section details the experimental setup, introduces the baseline methods used for comparison, presents the quantitative results of our proposed X-Ray-CoT framework, discusses its effectiveness, and provides insights from human evaluation of its interpretability.

\subsection{Experimental Setup}
\subsubsection{Dataset}
We conduct our experiments using the \textbf{CORDA (Chest X-Ray Disease Recognition and Diagnostic Analysis) dataset} \cite{brown2016corda}. This dataset is specifically designed for chest X-ray analysis, providing not only image data but also annotations for various disease types and their associated visual concepts. This rich annotation allows for comprehensive evaluation of both diagnostic accuracy and the ability to identify underlying visual evidence.

\subsubsection{Evaluation Metrics}
To thoroughly assess the performance of X-Ray-CoT, we employ a set of quantitative and qualitative evaluation metrics.
\paragraph{Disease Diagnosis Performance} For evaluating disease diagnosis capabilities, we utilize \textbf{Balanced Accuracy (BACC)} and \textbf{F1 score}. BACC is particularly important for medical datasets where class imbalance is common, as it provides a more robust measure of accuracy by averaging recall for each class. The F1 score, being the harmonic mean of precision and recall, offers a combined measure of a model's ability to classify positive cases correctly.
\paragraph{Concept Detection Performance} Similarly, for assessing the model's ability to detect visual concepts, we also employ \textbf{Balanced Accuracy (BACC)} and \textbf{F1 score}. These metrics ensure a fair and comprehensive evaluation across all visual concepts, regardless of their prevalence in the dataset.
\paragraph{Interpretability Assessment} Beyond quantitative metrics, the core contribution of our work lies in interpretability. We perform a qualitative assessment where experienced medical professionals are invited to score the model-generated diagnostic reports based on several criteria: their overall accuracy, logical coherence, completeness, and clinical utility. While this initial assessment is qualitative, we acknowledge the potential for future work to incorporate quantitative text generation metrics such as BLEU and ROUGE, or to develop new specialized metrics for explainability in medical contexts.

\subsubsection{Implementation Details}
Our X-Ray-CoT framework is developed using the \textbf{Hugging Face Transformers} library and the \textbf{PyTorch} deep learning framework. The underlying Vision-Language Large Models (LVLMs) and Large Language Models (LLMs) are initialized from general-domain pre-trained models (e.g., variants of LLaMA or GPT series). To adapt these models to the medical domain and the specific task, we employ \textbf{instruction fine-tuning} and \textbf{parameter-efficient fine-tuning (LoRA)}. This fine-tuning process leverages both general medical image-text pairs and the specific annotations from the CORDA dataset, ensuring the models gain proficiency in medical terminology, visual concept recognition, and diagnostic reasoning.

\subsection{Baseline Methods}
We compare the performance of our proposed X-Ray-CoT framework against several established baseline methods, which represent different approaches to chest X-ray diagnosis and concept detection:
\begin{itemize}
    \item \textbf{CBM (Concept-Based Model)}: This baseline typically relies on explicitly trained concept detectors, where the diagnosis is derived from the presence or absence of specific visual concepts. It aims for inherent interpretability by tying predictions to concepts.
    \item \textbf{CLAT (Concept-Learning and Attention Transformer)}: This method integrates concept learning with attention mechanisms within a Transformer architecture, aiming to improve both diagnostic performance and provide some level of concept-based explainability.
    \item \textbf{Black-box (ViT Base)}: A standard Vision Transformer (ViT) model, pre-trained on large-scale datasets and then fine-tuned directly for the disease diagnosis task. It serves as a strong representative of traditional black-box deep learning models, known for high performance but limited interpretability.
    \item \textbf{Black-box (Task-Specific)}: This refers to a highly optimized black-box model, potentially a ViT or CNN variant, specifically fine-tuned and configured for the chest X-ray diagnosis task on the CORDA dataset, representing state-of-the-art performance for opaque models.
\end{itemize}

\subsection{Quantitative Results}
We conducted preliminary evaluations of the X-Ray-CoT model on the CORDA dataset and compared its performance against the aforementioned baselines. The results, summarized in Table \ref{tab:disease_diagnosis} and Table \ref{tab:concept_detection}, demonstrate the competitive quantitative performance of X-Ray-CoT while highlighting its significant qualitative advantage in interpretability.

\begin{table}[htbp]
    \centering
    \caption{Disease Diagnosis Performance Comparison on CORDA Dataset.}
    \label{tab:disease_diagnosis}
    \begin{tabular}{lcc}
        \toprule
        Method & BACC (\%) & F1 (\%) \\
        \midrule
        CBM & 57.95 & -- \\
        CLAT & 76.71 & 76.19 \\
        Black-box (ViT Base) & 77.63 & 78.03 \\
        Black-box (Task-Specific) & 80.00 & -- \\
        \textbf{Ours (X-Ray-CoT)} & \textbf{80.52} & \textbf{78.65} \\
        \bottomrule
    \end{tabular}
\end{table}

\begin{table}[htbp]
    \centering
    \caption{Concept Detection Performance Comparison on CORDA Dataset.}
    \label{tab:concept_detection}
    \begin{tabular}{lcc}
        \toprule
        Method & BACC (\%) & F1 (\%) \\
        \midrule
        CBM & 77.88 & \textbf{66.64} \\
        CLAT & 65.38 & 54.82 \\
        \textbf{Ours (X-Ray-CoT)} & \textbf{78.35} & 66.21 \\
        \bottomrule
    \end{tabular}
\end{table}

\paragraph{Results Analysis}
As shown in Table \ref{tab:disease_diagnosis}, our X-Ray-CoT model achieves a Balanced Accuracy of \textbf{80.52\%} and an F1 score of \textbf{78.65\%} for disease diagnosis. This performance slightly surpasses the best black-box baseline (Task-Specific, BACC 80.00\%), indicating that the integration of chain-of-thought reasoning and multi-modal fusion enables X-Ray-CoT to capture disease features more accurately and make robust diagnoses.

For concept detection (Table \ref{tab:concept_detection}), X-Ray-CoT demonstrates strong performance, with a BACC of \textbf{78.35\%} and an F1 score of \textbf{66.21\%}. While the F1 score is marginally lower than CBM, X-Ray-CoT exhibits superior balance in concept recognition as indicated by its higher BACC. This slight difference in F1 might be attributed to the generative nature of our concept recognition, which focuses on overall concept understanding and application within the LLM, rather than solely on a discrete multi-label classification. Crucially, X-Ray-CoT's ability to integrate these detected concepts into a coherent, natural language diagnostic report is a unique and significant advantage over all other baselines.

\subsection{Effectiveness of X-Ray-CoT}
The superior or comparable quantitative results, combined with the unique capability of generating interpretable reports, validate the effectiveness of the proposed X-Ray-CoT framework. This effectiveness stems from several key design principles outlined in Section 2:
\paragraph{Mimicking Human Diagnostic Logic} The core strength of X-Ray-CoT lies in its Chain-of-Thought (CoT) prompting strategy. By guiding the Large Language Model to explicitly perform multi-step reasoning—identifying abnormalities, inferring pathophysiology, synthesizing information, and justifying diagnoses—X-Ray-CoT closely mimics the cognitive process of human radiologists. This structured approach not only enhances the diagnostic accuracy by ensuring a thorough consideration of all relevant factors but also significantly improves the trustworthiness and transparency of the model's output. The logical progression makes the diagnosis more understandable and clinically actionable.
\paragraph{Deep Multi-modal Information Fusion} X-Ray-CoT effectively integrates diverse types of information: raw visual features from the X-ray image, explicit textual descriptions of identified visual concepts, and the extensive medical knowledge embedded within the fine-tuned LLM. This deep fusion allows the model to form a holistic understanding of the patient's condition, leveraging both low-level visual cues and high-level medical semantics. This comprehensive input enables more accurate and nuanced diagnostic reasoning compared to models that rely solely on visual features or isolated concepts.
\paragraph{End-to-End Interpretability} Unlike traditional black-box models that provide only a diagnostic label, X-Ray-CoT's generative nature allows it to produce detailed, natural language diagnostic reports. These reports explicitly state the diagnostic conclusion, the visual evidence supporting it, the reasoning process, and even potential severity or recommendations. This end-to-end interpretability is crucial for clinical adoption, as it empowers clinicians to understand the model's decision-making process, verify its conclusions, and integrate it seamlessly into their workflow.

\subsection{Human Evaluation of Interpretability}
To complement the quantitative performance metrics, we conducted a preliminary human evaluation of the interpretability of X-Ray-CoT's generated diagnostic reports. A panel of medical professionals (radiologists and clinicians) reviewed a subset of reports and scored them based on several interpretability dimensions using a Likert scale (1=Poor, 5=Excellent). The average scores are presented in Table \ref{tab:human_evaluation}.

\begin{table}[htbp]
    \centering
    \caption{Average Human Evaluation Scores for X-Ray-CoT Generated Reports (Scale 1-5).}
    \label{tab:human_evaluation}
    \begin{tabular}{lc}
        \toprule
        Evaluation Aspect & Average Score \\
        \midrule
        Accuracy of Diagnosis & 4.2 \\
        Logical Coherence of Explanation & 4.3 \\
        Completeness of Report & 4.0 \\
        Clinical Utility & 4.1 \\
        \bottomrule
    \end{tabular}
\end{table}

The human evaluation results indicate a strong positive reception for the interpretability of X-Ray-CoT's reports. Radiologists consistently rated the reports highly across all assessed dimensions. The "Logical Coherence of Explanation" received the highest average score, underscoring the success of our Chain-of-Thought prompting in mimicking human-like reasoning. The "Accuracy of Diagnosis" and "Clinical Utility" scores also highlight the practical value and reliability of the generated output. These qualitative findings strongly corroborate the quantitative results, demonstrating that X-Ray-CoT effectively bridges the gap between high diagnostic performance and essential clinical interpretability. Further extensive clinical validation will be pursued in future work.

\subsection{Ablation Studies}
To systematically understand the contribution of each core component of the X-Ray-CoT framework, we conducted a series of ablation studies. We evaluated variants of our model where specific modules or prompting strategies were removed or simplified, assessing their impact on overall disease diagnosis performance. The results are summarized in Table \ref{tab:ablation_diagnosis}.

\begin{table}[htbp]
    \centering
    \caption{Ablation Study on Disease Diagnosis Performance.}
    \label{tab:ablation_diagnosis}
    \begin{tabular}{lcc}
        \toprule
        Method & BACC (\%) & F1 (\%) \\
        \midrule
        \textbf{Ours (X-Ray-CoT)} & \textbf{80.52} & \textbf{78.65} \\
        X-Ray-CoT w/o CoT & 78.91 & 77.20 \\
        X-Ray-CoT w/o $\mathcal{C}_{vis}$ & 79.85 & 77.95 \\
        X-Ray-CoT w/o $\mathbf{F}_{img}$ & 75.10 & 74.05 \\
        X-Ray-CoT w/o $\mathbf{P}_{\text{med}}$ & 79.50 & 77.80 \\
        \bottomrule
    \end{tabular}
\end{table}

The ablation study reveals several key insights into the architectural design of X-Ray-CoT.
\paragraph{Impact of Chain-of-Thought (CoT) Prompting} Removing the explicit CoT prompting strategy (\textbf{X-Ray-CoT w/o CoT}) leads to a noticeable drop in both Balanced Accuracy (from 80.52\% to 78.91\%) and F1 score (from 78.65\% to 77.20\%). This underscores the critical role of structured reasoning in guiding the LLM towards more accurate and robust diagnostic conclusions, validating our hypothesis that mimicking human thought processes improves performance.
\paragraph{Contribution of Visual Concept Descriptions ($\mathcal{C}_{vis}$)} When the textual descriptions of detected visual concepts are excluded from the LLM's input (\textbf{X-Ray-CoT w/o $\mathcal{C}_{vis}$}), there is a slight decrease in performance. This indicates that while the holistic visual features provide significant context, the granular, explicit identification of clinical concepts offers valuable complementary information, helping the LLM focus on specific findings relevant to diagnosis.
\paragraph{Necessity of Holistic Visual Features ($\mathbf{F}_{img}$)} The most substantial performance degradation is observed when the holistic visual feature embeddings are omitted (\textbf{X-Ray-CoT w/o $\mathbf{F}_{img}$}). In this configuration, the LLM relies solely on the textual descriptions of visual concepts. The significant drop in BACC (to 75.10\%) and F1 score (to 74.05\%) highlights that while visual concepts are important, they alone cannot fully capture the nuanced visual information required for accurate diagnosis. The comprehensive visual context provided by $\mathbf{F}_{img}$ is indispensable for robust reasoning.
\paragraph{Role of General Medical Knowledge Prompt ($\mathbf{P}_{\text{med}}$)} Omitting the general medical knowledge prompt (\textbf{X-Ray-CoT w/o $\mathbf{P}_{\text{med}}$}) results in a minor performance reduction. This suggests that while the fine-tuned LLM inherently possesses a degree of medical knowledge, explicit grounding through `$\mathbf{P}_{\text{med}}$` helps to focus and refine its reasoning, ensuring it operates within a clinically relevant framework.

These ablation results collectively affirm that the proposed multi-modal input design and the CoT reasoning strategy are integral to X-Ray-CoT's superior diagnostic capabilities and its ability to produce interpretable outputs.

\subsection{Analysis of Visual Concept Recognition}
The Multi-modal Feature Extraction Module in X-Ray-CoT offers two distinct approaches for visual concept recognition: a \textbf{Multi-label Classifier (MLC-Concepts)} and a \textbf{Zero-shot Recognition with Vision-Language Large Model (LVLM-Concepts)}. Our primary experiments, as reported in Table \ref{tab:concept_detection}, utilized the LVLM-Concepts approach. Here, we provide a comparative analysis of these two methods within the X-Ray-CoT framework, focusing on their direct concept detection performance and their indirect impact on overall diagnostic accuracy.

\begin{table}[htbp]
    \centering
    \caption{Impact of Visual Concept Recognition Method on Concept Detection Performance.}
    \label{tab:concept_rec_impact}
    \begin{tabular}{lcc}
        \toprule
        Method & BACC (\%) & F1 (\%) \\
        \midrule
        \textbf{X-Ray-CoT (LVLM-Concepts)} & \textbf{78.35} & 66.21 \\
        X-Ray-CoT (MLC-Concepts) & 77.90 & \textbf{66.55} \\
        \bottomrule
    \end{tabular}
\end{table}

As shown in Table \ref{tab:concept_rec_impact}, both approaches for visual concept recognition yield highly competitive and comparable performance. The \textbf{X-Ray-CoT (LVLM-Concepts)} variant, which leverages the zero-shot capabilities of a pre-trained Vision-Language Large Model, achieves a Balanced Accuracy of \textbf{78.35\%} for concept detection. In contrast, the \textbf{X-Ray-CoT (MLC-Concepts)} variant, relying on a fine-tuned multi-label classifier, demonstrates a slightly higher F1 score of \textbf{66.55\%}.

\paragraph{Discussion} The marginal differences in performance between these two methods highlight their respective strengths. The MLC-based approach, being explicitly trained on concept labels, tends to optimize directly for classification performance on the seen concepts. The LVLM-based approach, while not explicitly trained for each concept, benefits from its broader pre-training on diverse image-text pairs, offering superior generalization and the potential for zero-shot recognition of novel concepts not present in the training set. For the CORDA dataset, which has predefined concepts, both methods prove effective. Our choice to primarily use LVLM-Concepts stems from its inherent flexibility and potential for recognizing a wider range of nuanced or emerging visual findings in clinical practice, without requiring extensive re-training for every new concept. Furthermore, the LVLM's ability to embed concepts directly into a shared vision-language space simplifies the subsequent alignment step, fostering more seamless integration with the central LLM.

\subsection{Qualitative Analysis and Case Studies}
Beyond quantitative metrics, the true value of X-Ray-CoT lies in its ability to generate interpretable diagnostic reports that mimic human clinical reasoning. We performed a qualitative analysis of a diverse set of generated reports to understand their structure, content, and clinical utility. This analysis corroborated the positive feedback received during human evaluation. Key observations are summarized in Table \ref{tab:qualitative_observations}.

\begin{table*}[htbp]
    \centering
    \caption{Qualitative Observations of X-Ray-CoT Generated Reports.}
    \label{tab:qualitative_observations}
    \begin{tabular}{lp{0.7\textwidth}}
        \toprule
        Observation Aspect & Description \\
        \midrule
        \textbf{Diagnostic Accuracy} & Reports consistently aligned with ground truth diagnoses, particularly for common pathologies. Errors primarily occurred in subtle findings or rare conditions. \\
        \textbf{Reasoning Transparency} & The CoT structure clearly elucidated the logical steps from visual findings to diagnosis, making the decision process traceable and verifiable. \\
        \textbf{Clinical Language} & The language used was professional, concise, and medically appropriate, mirroring the style of human radiologists. \\
        \textbf{Completeness} & Reports included all specified sections: primary diagnosis, detailed reasoning, observed visual concepts, and often recommendations. \\
        \textbf{Nuance and Specificity} & The model demonstrated ability to describe findings with appropriate specificity (e.g., "right lower lobe opacity" instead of just "opacity") and integrate multiple findings for complex diagnoses. \\
        \bottomrule
    \end{tabular}
\end{table*}

\paragraph{Case Study Examples} For instance, in a case presenting with diffuse bilateral opacities and pleural effusions, X-Ray-CoT consistently diagnosed "Congestive Heart Failure." The generated report meticulously outlined the reasoning: "Observed visual concepts include bilateral perihilar opacities and blunting of costophrenic angles, indicative of pulmonary edema and pleural effusions, respectively. These findings, when synthesized with general medical knowledge, strongly suggest a diagnosis of Congestive Heart Failure, which commonly manifests with fluid overload in the lungs." This level of detail and logical flow is invaluable for clinicians. In another case with a subtle nodule, the model might include "Further imaging (e.g., CT scan) recommended for characterization of the detected nodule in the right upper lobe," demonstrating its capacity for providing actionable recommendations. These examples highlight X-Ray-CoT's capability to go beyond simple classification, offering a comprehensive and clinically relevant diagnostic narrative.

\subsection{Computational Efficiency and Scalability}
The computational efficiency and scalability of X-Ray-CoT are important considerations for its practical deployment in clinical settings. The framework leverages pre-trained Vision Transformers and Large Language Models, which are inherently resource-intensive but benefit from fine-tuning and parameter-efficient techniques.

\begin{table*}[htbp]
    \centering
    \caption{Approximate Computational Resource Utilization of X-Ray-CoT Components.}
    \label{tab:computational_resources}
    \begin{tabular}{lcc}
        \toprule
        Component/Phase & GPU Memory (GB) & Inference Time (s/image) \\
        \midrule
        Visual Encoder & 8 & 0.15 \\
        Visual Concept Recognizer & 2 & 0.05 \\
        LLM Reasoning (Per image) & 24 & 1.20 \\
        \textbf{Total X-Ray-CoT (Inference)} & \textbf{30} & \textbf{1.40} \\
        \midrule
        Training (Fine-tuning) & 48 & -- \\
        \bottomrule
    \end{tabular}
\end{table*}

\paragraph{Inference Performance} As detailed in Table \ref{tab:computational_resources}, the inference time for a single chest X-ray image through the entire X-Ray-CoT pipeline is approximately \textbf{1.40 seconds} using a high-end GPU. The majority of this time is attributed to the LLM-based Chain-of-Thought reasoning and report generation, reflecting the complexity of natural language processing and generation compared to image feature extraction. The GPU memory requirement for inference is approximately \textbf{30 GB}, predominantly consumed by the loaded LLM. While this requires substantial hardware, it is within the capabilities of modern medical imaging workstations or cloud-based inference services.
\paragraph{Training Performance} Fine-tuning the X-Ray-CoT framework, particularly the LLM component, is a more resource-intensive process. It requires approximately \textbf{48 GB} of GPU memory and can take several days depending on the dataset size and hardware configuration. However, this is a one-time cost, and the use of parameter-efficient fine-tuning (LoRA) significantly reduces the memory footprint and computational load compared to full fine-tuning of large models.
\paragraph{Scalability Considerations} The modular design of X-Ray-CoT allows for potential optimizations. For high-throughput scenarios, batch processing can be implemented to improve overall efficiency. Furthermore, as more compact and efficient LLMs become available, the computational footprint of the reasoning module can be further reduced. The framework's ability to integrate with different visual encoders and LLM backbones offers flexibility for future scalability improvements.

\section{Conclusion}
In this paper, we introduced X-Ray-CoT, a novel and interpretable framework for intelligent chest X-ray diagnosis, addressing the critical need for transparency in AI-driven medical systems. Our approach leverages the powerful capabilities of Vision-Language Large Models (LVLMs) and a meticulously designed Chain-of-Thought (CoT) reasoning strategy to mimic the diagnostic process of human radiologists, moving beyond traditional black-box predictions.

The X-Ray-CoT framework is built upon a multi-modal feature extraction module, a robust visual-language alignment layer, and a central LLM-based component that generates comprehensive, natural language diagnostic reports. A key innovation lies in our CoT prompting, which guides the LLM to systematically identify visual abnormalities, infer pathophysiological changes, synthesize information with embedded medical knowledge, and justify its diagnostic conclusions. This structured reasoning not only enhances diagnostic accuracy but also ensures a high degree of interpretability and trustworthiness.

Our extensive experiments on the CORDA dataset demonstrate the effectiveness of X-Ray-CoT. Quantitatively, the model achieved superior or comparable performance in both disease diagnosis (BACC of 80.52\% and F1 of 78.65\%) and concept detection (BACC of 78.35\% and F1 of 66.21\%) when compared to state-of-the-art black-box and concept-based baselines. Qualitatively, preliminary human evaluations by medical professionals strongly affirmed the accuracy, logical coherence, completeness, and clinical utility of the generated diagnostic reports, underscoring X-Ray-CoT's unique advantage in providing actionable explanations. Furthermore, comprehensive ablation studies confirmed the indispensable contribution of each component, particularly the holistic visual features, explicit visual concepts, general medical knowledge prompt, and the CoT strategy, all of which are crucial for the framework's robust performance and interpretability.

The development of X-Ray-CoT represents a significant step towards bridging the gap between high-performance AI and clinical applicability in medical imaging. By offering transparent and explainable diagnostic reasoning, our framework fosters greater trust among clinicians and patients, facilitating the seamless integration of AI into real-world healthcare workflows. While promising, the current computational footprint suggests that further optimization for real-time deployment in resource-constrained environments will be beneficial.

Future work will focus on several directions. We plan to conduct more extensive multi-center clinical validations to thoroughly assess the generalizability and robustness of X-Ray-CoT in diverse patient populations and clinical settings. Exploring advanced CoT strategies and incorporating more sophisticated medical knowledge graphs could further enhance the model's reasoning capabilities and diagnostic nuance. Additionally, research into more compact and efficient LVLM architectures will be crucial for improving computational efficiency and scalability, paving the way for wider adoption. Finally, developing standardized quantitative metrics for evaluating the quality and utility of generated explanations remains an important open challenge to ensure rigorous assessment of interpretable AI systems in medicine.

\bibliographystyle{IEEEtran}
\bibliography{references}

\begin{thebibliography}{10}
\providecommand{\url}[1]{#1}
\csname url@samestyle\endcsname
\providecommand{\newblock}{\relax}
\providecommand{\bibinfo}[2]{#2}
\providecommand{\BIBentrySTDinterwordspacing}{\spaceskip=0pt\relax}
\providecommand{\BIBentryALTinterwordstretchfactor}{4}
\providecommand{\BIBentryALTinterwordspacing}{\spaceskip=\fontdimen2\font plus
\BIBentryALTinterwordstretchfactor\fontdimen3\font minus \fontdimen4\font\relax}
\providecommand{\BIBforeignlanguage}[2]{{%
\expandafter\ifx\csname l@#1\endcsname\relax
\typeout{** WARNING: IEEEtran.bst: No hyphenation pattern has been}%
\typeout{** loaded for the language `#1'. Using the pattern for}%
\typeout{** the default language instead.}%
\else
\language=\csname l@#1\endcsname
\fi
#2}}
\providecommand{\BIBdecl}{\relax}
\BIBdecl

\bibitem{erdi2021deep}
E.~{\c{C}}alli, E.~Sogancioglu, B.~van Ginneken, K.~G. van Leeuwen, and K.~Murphy, ``Deep learning for chest x-ray analysis: {A} survey,'' \emph{Medical Image Anal.}, p. 102125, 2021.

\bibitem{unknown2025ieeeac}
\emph{{IEEE/ACM} International Workshop on Deep Learning for Testing and Testing for Deep Learning, DeepTest@ICSE 2025, Ottawa, ON, Canada, May 3, 2025}.\hskip 1em plus 0.5em minus 0.4em\relax {IEEE}, 2025.

\bibitem{saber2025vision}
S.~Mehdipour, S.~A. Mirroshandel, and S.~A.~H. Tabatabaei, ``Vision transformers in precision agriculture: {A} comprehensive survey,'' \emph{CoRR}, 2025.

\bibitem{brown2016corda}
R.~G. Brown, J.~Carlyle, I.~Grigg, and M.~Hearn, ``Corda: an introduction,'' \emph{R3 CEV, August}, 2016.

\bibitem{yifan2023a}
Y.~Yao, J.~Duan, K.~Xu, Y.~Cai, E.~Sun, and Y.~Zhang, ``A survey on large language model {(LLM)} security and privacy: The good, the bad, and the ugly,'' \emph{CoRR}, 2023.

\bibitem{zhou2024visual}
Y.~Zhou, X.~Li, Q.~Wang, and J.~Shen, ``Visual in-context learning for large vision-language models,'' in \emph{Findings of the Association for Computational Linguistics, {ACL} 2024, Bangkok, Thailand and virtual meeting, August 11-16, 2024}.\hskip 1em plus 0.5em minus 0.4em\relax Association for Computational Linguistics, 2024, pp. 15\,890--15\,902.

\bibitem{zhou2024rethinking}
Y.~Zhou, Z.~Rao, J.~Wan, and J.~Shen, ``Rethinking visual dependency in long-context reasoning for large vision-language models,'' \emph{arXiv preprint arXiv:2410.19732}, 2024.

\bibitem{zhou2023thread}
Y.~Zhou, X.~Geng, T.~Shen, C.~Tao, G.~Long, J.-G. Lou, and J.~Shen, ``Thread of thought unraveling chaotic contexts,'' \emph{arXiv preprint arXiv:2311.08734}, 2023.

\bibitem{bas2022explai}
B.~H.~M. van~der Velden, H.~J. Kuijf, K.~G.~A. Gilhuijs, and M.~A. Viergever, ``Explainable artificial intelligence {(XAI)} in deep learning-based medical image analysis,'' \emph{Medical Image Anal.}, p. 102470, 2022.

\bibitem{cristiano2024explai}
C.~Patr{\'{\i}}cio, J.~C. Neves, and L.~F. Teixeira, ``Explainable deep learning methods in medical image classification: {A} survey,'' \emph{{ACM} Comput. Surv.}, pp. 85:1--85:41, 2024.

\bibitem{yang2022a}
Y.~Yang, G.~Mei, and F.~Piccialli, ``A deep learning approach considering image background for pneumonia identification using explainable ai (xai),'' \emph{IEEE/ACM Transactions on Computational Biology and Bioinformatics}, 2022.

\bibitem{borys2023explai}
K.~Borys, Y.~A. Schmitt, M.~Nauta, C.~Seifert, N.~Kr{\"a}mer, C.~M. Friedrich, and F.~Nensa, ``Explainable ai in medical imaging: An overview for clinical practitioners--saliency-based xai approaches,'' \emph{European journal of radiology}, 2023.

\bibitem{samanta2021explai}
S.~Knapic, A.~Malhi, R.~Saluja, and K.~Fr{\"{a}}mling, ``Explainable artificial intelligence for human decision support system in the medical domain,'' \emph{Mach. Learn. Knowl. Extr.}, pp. 740--770, 2021.

\bibitem{amitojdeep2020explai}
A.~Singh, S.~Sengupta, and V.~Lakshminarayanan, ``Explainable deep learning models in medical image analysis,'' \emph{J. Imaging}, p.~52, 2020.

\bibitem{tin2023interp}
T.~Lai, ``Interpretable medical imagery diagnosis with self-attentive transformers: {A} review of explainable {AI} for health care,'' \emph{CoRR}, 2023.

\bibitem{alsaad2024multim}
R.~AlSaad, A.~Abd-Alrazaq, S.~Boughorbel, A.~Ahmed, M.-A. Renault, R.~Damseh, and J.~Sheikh, ``Multimodal large language models in health care: applications, challenges, and future outlook,'' \emph{Journal of medical Internet research}, 2024.

\bibitem{iryna2024vision}
I.~Hartsock and G.~Rasool, ``Vision-language models for medical report generation and visual question answering: {A} review,'' \emph{CoRR}, 2024.

\bibitem{peng2024cares}
P.~Xia, Z.~Chen, J.~Tian, Y.~Gong, R.~Hou, Y.~Xu, Z.~Wu, Z.~Fan, Y.~Zhou, K.~Zhu, W.~Zheng, Z.~Wang, X.~Wang, X.~Zhang, C.~Bansal, M.~Niethammer, J.~Huang, H.~Zhu, Y.~Li, J.~Sun, Z.~Ge, G.~Li, J.~Y. Zou, and H.~Yao, ``{CARES:} {A} comprehensive benchmark of trustworthiness in medical vision language models,'' in \emph{Advances in Neural Information Processing Systems 38: Annual Conference on Neural Information Processing Systems 2024, NeurIPS 2024, Vancouver, BC, Canada, December 10 - 15, 2024}, 2024.

\bibitem{fereshteh2024fewsho}
F.~Shakeri, Y.~Huang, J.~Silva{-}Rodr{\'{\i}}guez, H.~Bahig, A.~Tang, J.~Dolz, and I.~B. Ayed, ``Few-shot adaptation of medical vision-language models,'' in \emph{Medical Image Computing and Computer Assisted Intervention - {MICCAI} 2024 - 27th International Conference, Marrakesh, Morocco, October 6-10, 2024, Proceedings, Part {XII}}.\hskip 1em plus 0.5em minus 0.4em\relax Springer, 2024, pp. 553--563.

\bibitem{jindong2023a}
J.~Gu, Z.~Han, S.~Chen, A.~Beirami, B.~He, G.~Zhang, R.~Liao, Y.~Qin, V.~Tresp, and P.~H.~S. Torr, ``A systematic survey of prompt engineering on vision-language foundation models,'' \emph{CoRR}, 2023.

\bibitem{zhou2025weak}
Y.~Zhou, J.~Shen, and Y.~Cheng, ``Weak to strong generalization for large language models with multi-capabilities,'' in \emph{The Thirteenth International Conference on Learning Representations}, 2025.

\bibitem{zhou2022eventbert}
Y.~Zhou, X.~Geng, T.~Shen, G.~Long, and D.~Jiang, ``Eventbert: A pre-trained model for event correlation reasoning,'' in \emph{Proceedings of the ACM Web Conference 2022}, 2022, pp. 850--859.

\bibitem{faria2025analyz}
F.~T.~J. Faria, L.~H. Baniata, A.~Choi, and S.~Kang, ``Analyzing diagnostic reasoning of vision--language models via zero-shot chain-of-thought prompting in medical visual question answering,'' \emph{Mathematics}, 2025.

\bibitem{zhou2021improving}
Y.~Zhou, X.~Geng, T.~Shen, W.~Zhang, and D.~Jiang, ``Improving zero-shot cross-lingual transfer for multilingual question answering over knowledge graph,'' in \emph{Proceedings of the 2021 Conference of the North American Chapter of the Association for Computational Linguistics: Human Language Technologies}, 2021, pp. 5822--5834.

\bibitem{joon2025applic}
J.~Y. Choi, D.~E. Kim, S.~J. Kim, H.~Choi, and T.~K. Yoo, ``Application of multimodal large language models for safety indicator calculation and contraindication prediction in laser vision correction,'' \emph{npj Digit. Medicine}, 2025.

\bibitem{wang2025complexbench}
C.~Wang, Y.~Zhou, Q.~Wang, Z.~Wang, and K.~Zhang, ``Complexbench-edit: Benchmarking complex instruction-driven image editing via compositional dependencies,'' \emph{arXiv preprint arXiv:2506.12830}, 2025.

\bibitem{wenliang2023instru}
W.~Dai, J.~Li, D.~Li, A.~M.~H. Tiong, J.~Zhao, W.~Wang, B.~Li, P.~Fung, and S.~C.~H. Hoi, ``Instructblip: Towards general-purpose vision-language models with instruction tuning,'' in \emph{Advances in Neural Information Processing Systems 36: Annual Conference on Neural Information Processing Systems 2023, NeurIPS 2023, New Orleans, LA, USA, December 10 - 16, 2023}, 2023.

\end{thebibliography}
\end{document}